\documentclass[runningheads]{llncs}
\usepackage[T1]{fontenc}
\usepackage{graphicx,verbatim}
\usepackage{hyperref}
\usepackage{color}

\usepackage{orcidlink} 

\usepackage[export]{adjustbox} 
 \usepackage{float} 

\usepackage{arydshln}
 
\begin{document}
\title{Sex-based Bias Inherent in the Dice Similarity Coefficient: A Model Independent Analysis for Multiple Anatomical Structures}
\titlerunning{Sex-based Bias Inherent in the Dice Similarity Coefficient}
%

\iftrue
\author{Hartmut Häntze\inst{1,2,3}\orcidlink{0000-0002-6204-571X} \and
Myrthe Buser\inst{2}\orcidlink{0000-0003-0640-6434} \and
Alessa Hering\inst{2}\orcidlink{0000-0002-7602-803X} \and
Lisa C. Adams\inst{3}\orcidlink{0000-0001-5836-4542} \and
Keno K. Bressem\inst{3}\orcidlink{0000-0001-9249-8624}}

\authorrunning{H. Häntze et al.}
\institute{Charité - Universitätsmedizin Berlin, 12203 Berlin, Germany 
\email{hartmut.haentze@charite.de} \and
Radboudumc, 6525 GA Nijmegen, Netherlands \and 
Klinikum rechts der Isar, Technical University of Munich, 81675 Munich, Germany}

\else
\author{Anonymized Authors}
\authorrunning{Anonymized Author et al.}
\institute{Anonymized Affiliations  \\
\email{email@anonymized.com}}
\fi

\maketitle              
%
\begin{abstract}
Overlap-based metrics such as the Dice Similarity Coefficient (DSC) penalize segmentation errors more heavily in smaller structures. As organ size differs by sex, this implies that a segmentation error of equal magnitude may result in lower DSCs in women due to their smaller average organ volumes compared to men. While previous work has examined sex-based differences in models or datasets, no study has yet investigated the potential bias introduced by the DSC itself. This study quantifies sex-based differences of the DSC and the normalized DSC in an idealized setting independent of specific models. We applied equally-sized synthetic errors to manual MRI annotations from 50 participants to ensure sex-based comparability. Even minimal errors (e.g., a 1 mm boundary shift) produced systematic DSC differences between sexes. For small structures, average DSC differences were around 0.03; for medium-sized structures around 0.01. Only large structures (i.e., lungs and liver) were mostly unaffected, with sex-based DSC differences close to zero. These findings underline that fairness studies using the DSC as an evaluation metric should not expect identical scores between men and women, as the metric itself introduces bias. A segmentation model may perform equally well across sexes in terms of error magnitude, even if observed DSC values suggest otherwise. Importantly, our work raises awareness of a previously underexplored source of sex-based differences in segmentation performance. One that arises not from model behavior, but from the metric itself. Recognizing this factor is essential for more accurate and fair evaluations in medical image analysis.

\keywords{Fairness \and Segmentation \and Dice Similarity Coefficient.}
\end{abstract}
\section{Introduction}
The Dice Similarity Coefficient (DSC) is the most commonly used segmentation metric in medical imaging. Together with Intersection over Union it is the default recommendation of frameworks such as Metrics Reloaded \cite{maier2024metrics}. Its pitfalls, such as indifference to multiple classes \cite{reinke2021common} or size-dependent performance \cite{dice1945measures,reinke2021common} are well reported in the literature and, dependent on the task, it can be advisable to pair it with other non-overlap metrics, such as Hausdorff Distance or Normalized Surface Distance \cite{maier2024metrics}. Alternatively, volume-corrected metrics such as the normalized DSC (nDSC) have been proposed and demonstrate reduced bias in contexts with substantial target volume variation, such as white matter lesion segmentation \cite{raina2023tackling}, although they have yet to see widespread adoption.

The DSC has been widely used to evaluate the fairness of segmentation models.  
Typically, models are trained on mixed-sex datasets, and DSCs are calculated separately for male and female subgroups. A common assumption is that a fair model should yield comparable DSCs across sexes, which can be controlled by statistical tests.
Some studies report higher DSCs for men \cite{hantze2024mrsegmentator,ioannou2022study}, women \cite{afzal2023towards}, or equal outcomes for both sexes \cite{de2025robust,lee2022systematic,pettit2022nnu,puyol2022fairness}.
When a performance gap is observed, it can be attributed to unbalanced training data or anatomical differences that make one sex more difficult to segment. However, a third possibility is frequently overlooked: the DSC itself may introduce bias related to sex-specific organ size.

\begin{figure}
\centering
\includegraphics[width=0.7\textwidth]{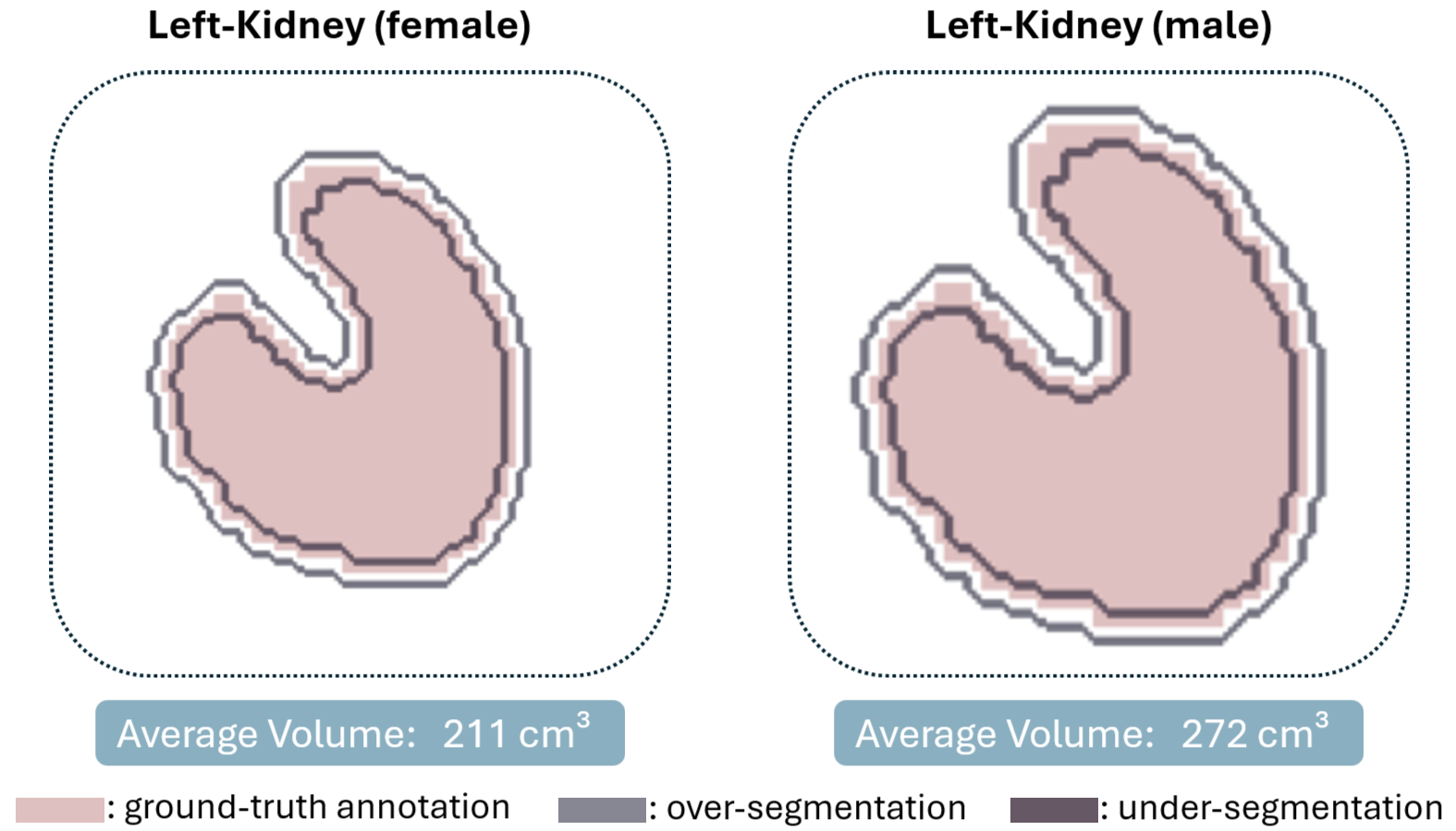}
\caption{Example annotation-mask of a left kidney with simulated over- and under-segmentation using a uniform 3 mm margin (grey and black outlines). In our cohort, female kidneys are on average smaller than male ones. Applying the same fixed error systematically across all kidneys in the cohort may result in different DSC values despite an identical error margin, as the increased size of male kidneys positively influences the outcome of the DSC calculation.} \label{fig1}
\end{figure}

It is well documented that the DSC disproportionately penalizes errors in smaller structures \cite{reinke2021common}, yet few discuss the implications this has on comparisons of subgroups with different organ volumes. Ferrante and Echeveste \cite{ferrante2025open} caution against directly comparing DSC values between pediatric and adult patients for this reason and explicitly raise concerns about sex-based comparisons, given known volume differences between male and female organs \cite{geraghty2004normal}. 
Despite these known limitations of the DSC, their specific impact on sex-based analyses has not been systematically quantified. For example, Häntze et al. \cite{hantze2024mrsegmentator} reported average DSC scores of 0.89 for men and 0.87 for women, but it remains unclear whether this reflects true model performance differences or biases inherent to the metric itself. 

In this study, we explicitly quantify sex-related volume bias in the DSC. Unlike previous work, we do not evaluate the output of a specific segmentation model. Instead, to ensure comparability, we simulate uniform over- and under-segmentation by adding or removing a fixed voxel margin along structure boundaries (Fig. \ref{fig1}). This approach allows us to precisely control the error magnitude and apply identical modifications to subjects of both sexes. Consequently, any observed differences in evaluation metrics cannot be attributed to model behavior or training data composition but must stem from subject-specific characteristics such as sex and organ volume.

\section{Methods and Materials}

\subsection{Study Design}
This study is a retrospective simulation. Explicit ethical consent was not required, as the analysis was performed on data from the German National Cohort \cite{nako2,nako1} accessed through application number 836.

\subsection{Dataset Description}
We used a dataset of whole-body in-phase gradient echo sequences of 50 participants (25 male, 25 female) from the German National Cohort with similar age range and BMI (Table \ref{tab1}). The dataset includes annotations for 40 anatomical structures that were quality-controlled by two board-certified radiologists as described in \cite{hantze2024mrsegmentator}. We excluded left and right femur annotations, as these structures were not fully captured in all sequences. We further removed one annotation of the left kidney that we identified as a pathologic outlier. Additionally, six participants were missing a gallbladder, either due to cholecystectomy or annotation mistakes. A complete list of structures, along with their average volumes, is provided in Table \ref{tab2}. Based on these volumes we categorized the structures into three groups: small (volume < 100 cm³), medium (100 to 1,000 cm³) and large (volume > 1,000 cm³). We chose these round number thresholds as they are easy to interpret and to apply consistently.

\begin{table}
\centering
\caption{Summary statistics (median with interquartile range in parentheses).}\label{tab1}
\begin{tabular}{|l|l|l|}
\hline
\textbf{Measure} & \textbf{Male (n=25)} & \textbf{Female (n=25)} \\
\hline
Age & 54.0 (14.0) & 52.0 (13.0) \\
Weight (kg) & 81.0 (17.0) & 64.0 (18.0) \\
Height (m) & 1.78 (0.07) & 1.64 (0.09) \\
BMI & 25.69 (5.05) & 24.14 (5.42) \\
\hline
\end{tabular}
\end{table}

\subsection{Evaluation Metrics}
We evaluated segmentation quality using the DSC, as it is the most frequently used metric in medical segmentation.
Additionally, we used the nDSC due to its proposed advantages; that is robustness against large volumetric variance. The nDSC achieves this by adjusting the DSC by the mean fraction of the positive class for the target structure and group. In this study, this adjustment was applied separately for male and female subjects using the mean volume of each anatomical structure by sex.
We did not test distance-based metrics such as normalized surface distance or Hausdorff distance. They are, by definition, not affected by volume, hence, errors that we induce with binary dilation will yield the same metrics.

\subsection{Experiments}
To assess how much the DSC and nDSC differentiate between men and women, we introduced controlled segmentation errors into our ground-truth masks. These synthetic errors are designed to resemble common failures of automated segmentation models, namely over-segmentation and under-segmentation. Because the DSC depends only on the overlap of two shapes, not on the spatial distribution of errors, we can simulate both under- and over-segmentation by applying binary dilation or erosion with a uniform margin around each reference structure, using the scikit-image \cite{van2014scikit} and Monai \cite{cardoso2022monai} python libraries.
We resampled the sequences to a isotopic spacing of 1 mm and conducted two experiments: First, to simulate a very good model with almost perfect segmentations we added an error margin for dilation and erosion of 1 mm. Second, to simulate a good but not perfect model, we added an error margin of 3 mm. Note, that due to the isotopic spacing 1 mm equals exactly the width of one voxel. 
During simulation, we applied both dilation and erosion independently to the reference masks of each participant and structure. We then computed the evaluation metrics for both error types and averaged the results to obtain a single score per structure and participant. 
To prevent adjacent structures from merging during dilation, all organs were processed independently during both the error simulation and evaluation steps.

\subsection{Statistical Analysis}
Since organ volumes differ between men and women, and DSC values are inherently linked to organ size, we expect to find corresponding DSC differences, as this relationship can be derived analytically. Consequently, this article's aim is not to only test whether differences exist, but rather to quantify how large these differences are. For this, we focused on the magnitude of the effects, reporting mean DSC and nDSC differences between male and female participants along with 95\% confidence intervals. We performed this analysis for each anatomical structure, across the three volume-based groups, and for both error margins (1 and 3 mm). 
Within each volume group, we assessed statistical significance using Welch's t-test and controlled the false discovery rate using the Benjamini-Hochberg correction.

\section{Results}
Applying the 1 mm error margin to the annotations resulted in average DSC values from 0.97 $\pm$ 0.00 (both lungs, and liver), up to 0.72 $\pm$ 0.10 (right adrenal gland). Table \ref{tab2} shows the specific DSC values for each structure for this error margin. The 3 mm error margin further reduced the DSC values, which range from 0.92 $\pm$ 0.01 up to 0.35 $\pm$ 0.10, respectively.
The volume corrected nDSC values were generally larger than their DSC counterparts; ranging from 0.98 $\pm$ 0.00 (lungs and liver) to 0.83 $\pm$ 0.07 (right adrenal gland) for 1 mm and from 0.95 $\pm$ 0.00 (right lung) to  0.58 $\pm$ 0.04  (left and right iliac arteries) for 3 mm.

\begin{figure}
\includegraphics[width=\textwidth]{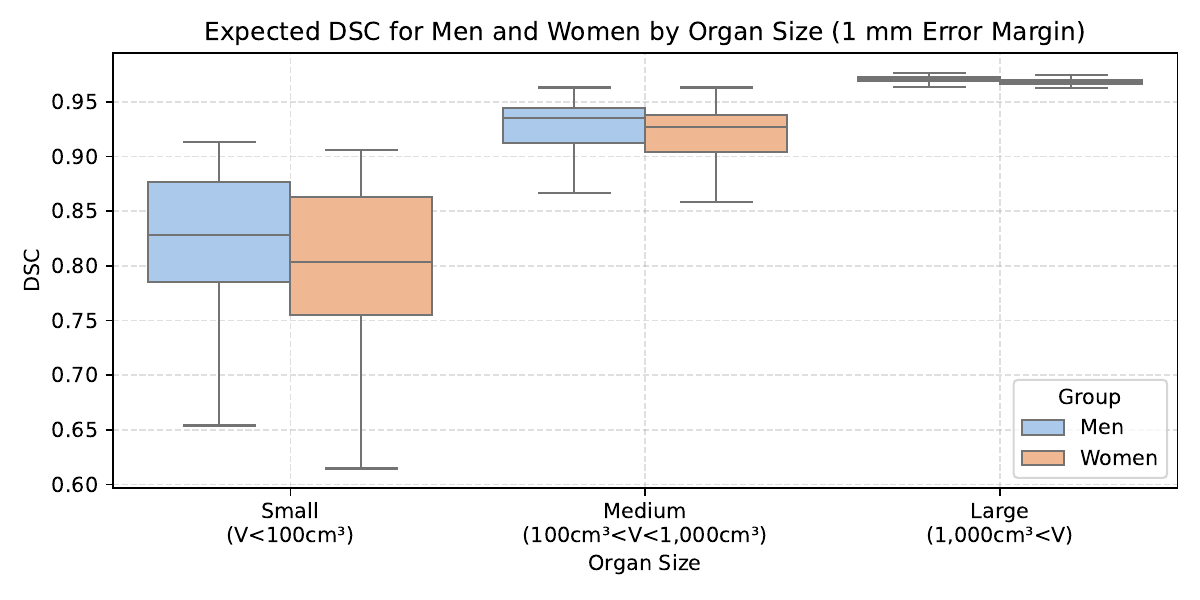}
\caption{Box-plot that shows the Dice similarity Coefficient (DSC) for an error margin of 1 mm grouped by sex and organ size. Smaller structures are proportionally more affected by volume differences, which results in a higher average DSC difference between men and women. Average differences are 0.03 for small structures, 0.01 for medium structures and 0.00 for large structures.} \label{fig2}
\end{figure}

The sex-stratified differences for the three volume categories are listed in Table \ref{tab3}. For the 1 mm error margin both sexes had average DSC differences of 0.00 for large, 0.01 for medium and 0.03 for small structures. The nDSC differences are 0.00, 0.00 and 0.04 respectively. The 3 mm margin further increased the differences for both metrics; all differences are significant with adjusted p < 0.001. The DSC value distributions by sex for the 1 mm margin are visualized in Fig. \ref{fig2}.

\begin{table}
\centering
\caption{Expected metric differences plus 95\% confidence intervals for segmentations with an equal error margin between men and women. All differences are significant with adjusted p < 0.001.}\label{tab3}
\begin{tabular}{|l|l|l|l|}
\hline
\textbf{Category} & \textbf{Error Margin} & \textbf{DSC} & \textbf{nDSC} \\
\hline
Large & 1 mm & 0.00 (0.00, 0.00)  & 0.00 (0.00, 0.00) \\
      & 3 mm & 0.01 (0.00, 0.01)  & 0.00 (0.00, 0.01)\\
Medium & 1 mm  & 0.01 (0.00, 0.01) & 0.01 (0.00, 0.01)\\
       & 3 mm  & 0.02 (0.01, 0.03)  & 0.02 (0.01, 0.02) \\
Small & 1 mm  & 0.03 (0.02, 0.04)  & 0.02 (0.01, 0.02) \\
      & 3 mm  & 0.06 (0.04, 0.08)  & 0.04 (0.03, 0.05) \\
\hline
\end{tabular}
\end{table}

\section{Discussion}
While multiple papers \cite{reinke2021common} and frameworks, such as metrics reloaded \cite{maier2024metrics}, mention the Dice Similarity Coefficient's (DSC) dependence on volume, none discuss the direct implication this has on sex: A subgroup with a larger average organ volume can be expected to have a higher DSC due to inherent bias in the metric itself, independent of model performance.

To quantify this bias we added synthetic segmentation errors to ground-truth annotations and calculated the resulting DSC. 
Our results demonstrate that even models with high overall accuracy and a small error margin of just 1 mm can produce systematically different DSC values between men and women. For small anatomical structures, we observed an average DSC difference of 0.03; for medium-sized structures, the difference was around 0.01. Only large structures (i.e., lungs and liver) were mostly unaffected by this bias, with DSC differences close to zero (while this difference was statistically significant, we consider it to be of limited practical relevance). 
When simulating a larger error margin of 3 mm, these discrepancies became more pronounced, with differences reaching up to 0.06 for small structures. In contrast, the normalized DSC (nDSC) was less sensitive towards sex-specific volume differences: for the 1 mm error margin nDSC differences were close to zero for both large and medium structures and reduced to 0.02 for small structures, compared to the standard DSC.

\begin{table}
\caption{\textbf{Average volumes and DSC values for all structures} Modifying the ground truth annotations with a uniform 1 mm error margin results in different Dice Similarity Coefficients (DSC) across structures. Column three reports the average DSC values (± standard deviation) across all participants. Column four shows the average DSC differences between men and women, along with the 95\% confidence intervals. Volume categories (large, medium, small) are separated by dashed lines.}
\label{tab2}
\begin{tabular}{lccc}
\textbf{Class} & \textbf{Avg. Volume } &  \textbf{ DSC (1mm) } & \textbf{ \(\Delta \)DSC (1mm) } \\
\hline
right lung & 1948 cm³ & 0.97 $\pm$ 0.00 & 0.00 [0.00, 0.00] \\
left lung & 1670 cm³ & 0.97 $\pm$ 0.00 & 0.00 [0.00, 0.00] \\
liver & 1575 cm³ & 0.97 $\pm$ 0.00 & 0.00 [0.00, 0.00] \\
\hdashline
small bowel & 723 cm³ & 0.92 $\pm$ 0.01 & 0.01 [0.00, 0.01] \\
colon & 650 cm³ & 0.92 $\pm$ 0.01 & 0.00 [-0.01, 0.01] \\
heart & 622 cm³ & 0.96 $\pm$ 0.00 & 0.00 [0.00, 0.01] \\
right gluteus maximus & 568 cm³ & 0.95 $\pm$ 0.01 & 0.01 [0.01, 0.01] \\
left gluteus maximus & 527 cm³ & 0.95 $\pm$ 0.01 & 0.01 [0.00, 0.01] \\
left autochthonous muscle & 418 cm³ & 0.94 $\pm$ 0.01 & 0.01 [0.01, 0.01] \\
right autochthonous muscle & 410 cm³ & 0.94 $\pm$ 0.01 & 0.01 [0.00, 0.01] \\
stomach & 350 cm³ & 0.94 $\pm$ 0.01 & 0.01 [0.00, 0.01] \\
spleen & 321 cm³ & 0.94 $\pm$ 0.01 & 0.00 [0.00, 0.01] \\
right kidney & 275 cm³ & 0.94 $\pm$ 0.01 & 0.00 [0.00, 0.01] \\
right hip & 262 cm³ & 0.90 $\pm$ 0.01 & 0.01 [0.01, 0.01] \\
left hip & 261 cm³ & 0.90 $\pm$ 0.01 & 0.01 [0.00, 0.01] \\
left iliopsoas muscle & 257 cm³ & 0.92 $\pm$ 0.01 & 0.02 [0.01, 0.02] \\
left kidney & 241 cm³ & 0.94 $\pm$ 0.01 & 0.01 [0.00, 0.01] \\
right iliopsoas muscle & 241 cm³ & 0.92 $\pm$ 0.01 & 0.02 [0.01, 0.02] \\
right gluteus medius & 227 cm³ & 0.94 $\pm$ 0.01 & 0.01 [0.01, 0.01] \\
spine & 221 cm³ & 0.89 $\pm$ 0.01 & 0.01 [0.01, 0.01] \\
aorta & 221 cm³ & 0.91 $\pm$ 0.01 & 0.01 [0.01, 0.01] \\
left gluteus medius & 211 cm³ & 0.94 $\pm$ 0.01 & 0.01 [0.00, 0.01] \\
urinary bladder & 191 cm³ & 0.94 $\pm$ 0.01 & 0.00 [-0.01, 0.01] \\
sacrum & 170 cm³ & 0.91 $\pm$ 0.01 & 0.01 [0.00, 0.01] \\
pancreas & 130 cm³ & 0.89 $\pm$ 0.01 & 0.01 [0.01, 0.02] \\
gallbladder & 123 cm³ & 0.85 $\pm$ 0.05 & 0.01 [-0.02, 0.04] \\
\hdashline
inferior vena cava & 91 cm³ & 0.88 $\pm$ 0.01 & 0.01 [0.01, 0.02] \\
right adrenal gland & 86 cm³ & 0.72 $\pm$ 0.10 & 0.04 [-0.02, 0.10] \\
left adrenal gland & 84 cm³ & 0.75 $\pm$ 0.06 & 0.05 [0.02, 0.08] \\
left iliac vena & 63 cm³ & 0.83 $\pm$ 0.02 & 0.02 [0.02, 0.03] \\
right iliac vena & 63 cm³ & 0.82 $\pm$ 0.02 & 0.03 [0.02, 0.04] \\
duodenum & 63 cm³ & 0.86 $\pm$ 0.03 & 0.03 [0.01, 0.04] \\
right gluteus minimus & 57 cm³ & 0.89 $\pm$ 0.01 & 0.01 [0.00, 0.01] \\
esophagus & 55 cm³ & 0.81 $\pm$ 0.02 & 0.02 [0.01, 0.03] \\
left gluteus minimus & 54 cm³ & 0.89 $\pm$ 0.01 & 0.01 [0.00, 0.01] \\
left iliac artery & 54 cm³ & 0.76 $\pm$ 0.03 & 0.04 [0.03, 0.06] \\
portal vein and splenic vein & 50 cm³ & 0.78 $\pm$ 0.02 & 0.01 [0.00, 0.03] \\
right iliac artery & 44 cm³ & 0.75 $\pm$ 0.04 & 0.05 [0.03, 0.06] \\

\end{tabular}
\end{table}

The choice of an appropriate error margin is closely tied to image resolution. For example, with a slice thickness of 3 mm, a 1 mm segmentation error is not feasible due to voxel size constraints. In such cases, the 3 mm error simulations more accurately reflect the impact of a one-voxel segmentation error and are therefore more applicable. Notably, the average DSC values from our 1 mm error simulations closely align with those reported for real-world models in the literature. For instance, Kart et al. \cite{kart2021deep} reported DSC values of 0.97 (liver), 0.95 (spleen), 0.95 (left kidney), 0.95 (right kidney), and 0.87 (pancreas) on the same cohort. Our simulations yielded comparable values: 0.97, 0.94, 0.94, 0.94, and 0.89, respectively.

It is important to understand that our results do not indicate a flaw in the DSC's calculation itself. The metric behaves as intended: the relative impact of a constant-sized error increases as the size of the target structure decreases; and the DSC correctly captures this. However, interpreting such differences as evidence of model unfairness can be misleading, as the absolute error may be identical for both groups. Even a segmentation model that operates in a sex-neutral manner can yield different DSC values between men and women due to inherent anatomical volume differences.

The findings of this study are expected to generalize to other 3D imaging modalities such as CT, as our analysis was based solely on annotations and did not incorporate MRI-specific features. For 2D modalities like X-ray, results are likely to differ as the dimensionality shift from three to two alters the voxel count per structure, and therefore also the calculation of the DSC.

This paper has limitations. Real-world segmentation errors are highly heterogeneous and can include not only over- and under-segmentation but also false positives and complete omissions of structures. By simulating errors through binary dilation and erosion, we simplify this complexity and inevitably lose some similarity to real model behavior. However, the purpose of this study is not to reproduce all error types but to quantify potential DSC differences between men and women under idealized conditions. 
Further, it should be noted that our analysis was conducted on a cohort of German participants, and results may not generalize to populations with different anatomical or sex-based characteristics.
Hence, the outcomes presented here should be interpreted as a rough estimate of the potential magnitude of sex-related differences of the DSC. To support comparability, we reported the average organ volumes for our cohort alongside the corresponding metrics (Table \ref{tab2}).

\section{Conclusion}
In this study, we quantified the expected DSC differences between men and women under the assumption of an ideal non-discriminatory segmentation model. Observed DSC differences vary in magnitude from 0.00 to 0.06. This shows that differences in DSC values between sexes do not necessarily indicate model bias. Rather, these discrepancies may be due to a volume-dependent and thus sex-dependent bias intrinsic to the DSC itself, particularly when evaluating small structures below 100 cm³.
Importantly, our results do not aim to replace the DSC, as multiple studies have clearly demonstrated its value in highlighting both negligible \cite{de2025robust,lee2022systematic,pettit2022nnu,puyol2022fairness} and substantial  \cite{afzal2023towards,ioannou2022study} sex-based differences. However, we show that systematic differences can arise even under optimal conditions. Consequently, when a segmentation model appears to underperform for one sex, especially for women, it is worth considering whether the metric itself contributes to the observed disparity.
Simulations like those presented in this article can help isolate metric-induced effects, and comparing results with alternative metrics such as distance-based metrics or the nDSC \cite{raina2023tackling} may offer a more complete assessment of fairness.
\bibliographystyle{splncs04}
\bibliography{main}

\begin{credits}
\subsubsection{\ackname}
This project was conducted with data from the German National Cohort (NAKO) (www.nako.de ). The NAKO is funded by the Federal Ministry of Education and Research (BMBF) [project funding reference numbers: 01ER1301A/B/C, 01ER1511D and 01ER1801A/B/C/D], federal states of Germany and the Helmholtz Association, the participating universities and the institutes of the Leibniz Association. We thank all participants who took part in the NAKO study and the staff of this research initiative. Much of the computation resources required for this research was performed on computational hardware generously provided by the Charité HPC cluster. Funded by the European Union. Views and opinions expressed are however those of the author(s) only and do not necessarily reflect those of the European Union or European Health and Digital Executive Agency (HADEA). Neither the European Union nor the granting authority can be held responsible for them.

\begin{figure}[H]
\includegraphics[width=0.3\linewidth, right]{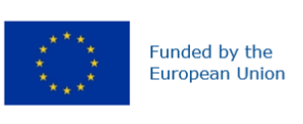}
\label{fig:eu_funding}
\end{figure}

\subsubsection{\discintname}
The authors have no competing interests to declare that are
relevant to the content of this article.

\end{credits}

\end{document}